\documentclass[10pt,twocolumn,letterpaper]{article}
\usepackage[pagenumbers]{cvpr}

\usepackage{cuted} 

\usepackage{currfile} 

\usepackage{caption} 
\usepackage{multirow}
\usepackage{colortbl}

\definecolor{cvprblue}{rgb}{0.21,0.49,0.74}
\usepackage[breaklinks,colorlinks,allcolors=cvprblue]{hyperref}
\hypersetup{
    pdftitle={SPGSR: A Small Pixel-Space Generator for One-Step Super-Resolution},
    pdfauthor={Yun Yue, Luming Liang, Yiqi Zhong}
}

\title{SPGSR: A Small Pixel-Space Generator for One-Step Super-Resolution}
\author{Yun Yue \qquad Luming Liang \qquad Yiqi Zhong\\Microsoft}

\begin{document}
\raggedbottom
\maketitle
\begin{strip}
\centering
\includegraphics[width=0.95\linewidth]{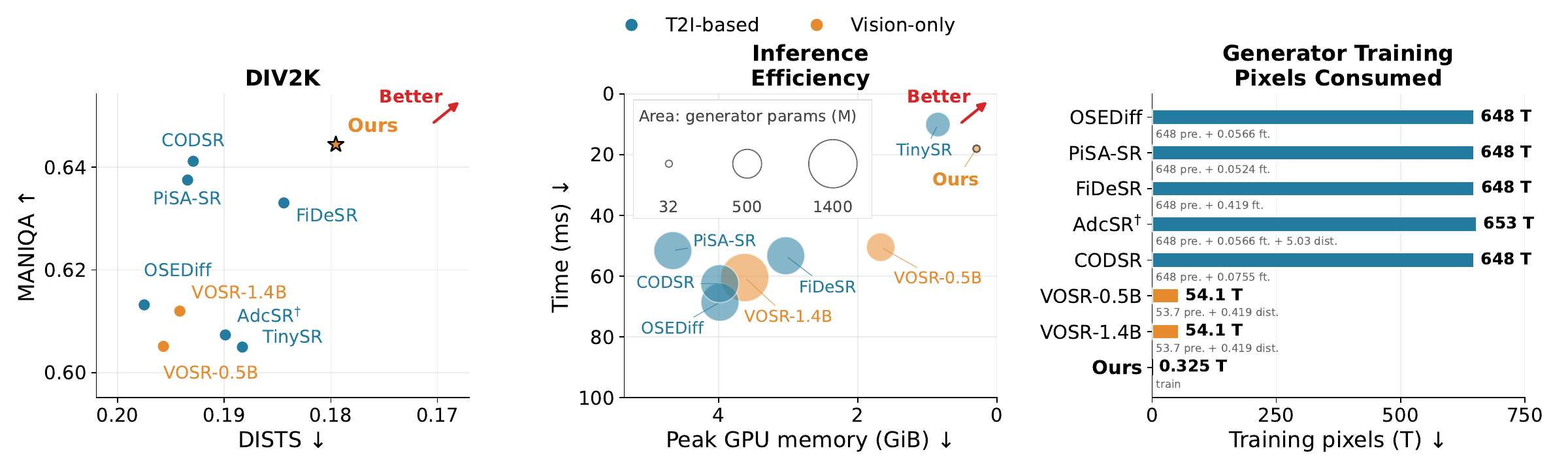}
\captionsetup{hypcap=false}
\captionof{figure}{\textbf{Quality, inference efficiency, and generator training budget.}
SPGSR achieves competitive restoration quality on DIV2K with a compact vision-only inference pipeline.
Panel 2 reports full GPU-pipeline latency and peak allocated memory under FP16 inference (RTX PRO 6000, batch 1, $512\times512$ outputs).
Bubble areas represent generator parameters.
$^{\dagger}$AdcSR is excluded from Panel 2 due to FP16 overflow; in matched FP32 inference, SPGSR uses 11.8\% of AdcSR's total inference parameters (53.86M vs.\ 456.07M) and achieves a $1.62\times$ GPU-pipeline speedup.}
\captionsetup{hypcap=true}
\label{fig:\currfilebase}%
\end{strip}
\begin{abstract}
In one-step super-resolution (SR), vision-only models extract semantic
cues directly from the low-resolution (LR) image, keeping those cues
spatially aligned with the structures they help restore.
Recent models achieve high restoration quality with generators of up to
1.4 billion parameters. Given how much structure the LR image already
provides, could a much smaller generator achieve comparable quality?
We introduce SPGSR, a vision-only framework that adds a learned RGB correction to the LR image.
A frozen DINOv2 guides prediction, and a lightweight block refines the correction locally.
SPGSR achieves higher PSNR/SSIM and lower LPIPS/DISTS than VOSR, a leading vision-only SR method, across three benchmarks.
Its $31.8\mathrm{M}$-parameter generator is just $6.5\%$ the size of VOSR-0.5B's.
SPGSR also achieves competitive quality against T2I-based SR methods.
SPGSR's generator uses only $0.60\%$ of the HR pixels used to train and distill VOSR's generator.
For $512\times512$ outputs, SPGSR achieves $2.80\times$ and $3.35\times$ GPU-pipeline speedups over VOSR-0.5B and VOSR-1.4B, respectively.
Code will be made publicly available.

\end{abstract}
\section{Introduction}
\label{sec:intro}

Real-world single-image super-resolution (SR) must recover plausible
high-resolution (HR) detail from a low-resolution (LR) image while
preserving the structure of the observed scene. Recent methods draw on
pretrained text-to-image (T2I) diffusion models to synthesize realistic
textures~\cite{wang2024exploiting,lin2024diffbir,wu2024one}.
In standard SR, however, the LR image is the only observation available
at inference. A text prompt inferred from it can help identify an object,
but offers limited guidance about where a contour should run or how a
texture should continue across a surface. Visual features can capture
that same semantic context while retaining its spatial correspondence
with the input. Paired LR and HR images therefore provide a natural way
to learn a vision-only restoration prior: they directly supervise which
fine details are consistent with the surviving visual cues.

VOSR~\cite{wu2026vosr} demonstrates that such a prior can achieve
competitive perceptual quality without T2I pretraining. It trains a
latent diffusion transformer from scratch, conditioning generation on
both the spatial structure and the semantic features of the LR image.
VOSR explores generator backbones with 0.5B and 1.4B parameters and employs distillation to enable one-step inference. While distillation reduces the number of sampling steps, each image still requires a forward pass through the full generator. Given how much scene structure is already visible in the LR image,
\textit{does high-quality one-step SR really require such a large generator?}

To explore this possibility, we introduce \textbf{SPGSR} (Small Pixel-Space
Generator for One-Step Super-Resolution), a compact
vision-only generator for one-step SR. We build on the visually guided
pixel-space transformer of PixRestore~\cite{sun2026pixrestore} and
adapt its generation process to the structure already present in the
LR observation. Specifically, SPGSR starts from the upsampled LR image
and predicts an RGB residual that is added back to the same image.
This gives the generator a direct path for retaining observed structure
while learning the corrections needed to remove artifacts and restore
missing detail.

To predict the RGB residual, SPGSR combines the visual context of the
LR image with local pixel information. Following
PixRestore~\cite{sun2026pixrestore}, we guide the transformer with
features from multiple layers of a frozen DINOv2
encoder~\cite{oquab2023dinov2}. During training, agreement between
paired LR and HR features teaches a router which layers to emphasize
for each input. Motivated by residual learning~\cite{he2016deep} and
its use in SR~\cite{zhang2018rcan,wang2018esrgan}, we introduce a
lightweight Pixel Residual Refinement Block (PixelRRB) to refine the
transformer's patchwise prediction.
FiDeSR~\cite{kim2026fidesr} applies residual refinement in latent
space before VAE decoding. PixelRRB operates directly on the predicted
RGB residual and the upsampled LR image, allowing neighboring pixels
on either side of a patch boundary to inform the final correction.

We train SPGSR from scratch in two stages. We first train the generator
to recover the HR target from linear interpolations between paired LR
and HR images, always predicting the residual relative to the LR image.
We then fine-tune it from the LR endpoint used at inference, adding
perceptual and DINO-based adversarial losses. This directly supervises
the one-step mapping with paired images and requires no separate SR
teacher. Across both stages, the 31.8M-parameter generator uses only
0.60\% of VOSR's combined generator pretraining and distillation
budget, measured in HR training pixels.
As shown in Fig.~\ref{fig:teaser}, SPGSR achieves competitive
restoration quality on DIV2K with a substantially smaller generator
and faster inference than VOSR-0.5B.

Comparisons with other compact SR methods, such as
AdcSR~\cite{chen2025adcsr}, TinySR~\cite{dong2025tinysr}, and
PocketSR~\cite{sun2025pocketsr}, demonstrate SPGSR's competitive
performance and favorable trade-off between restoration quality and
model size.

Our contributions are threefold.
\begin{itemize}
    \item We introduce \textbf{SPGSR}, a compact vision-only generator
    that performs one-step SR by predicting an RGB residual from the
    upsampled LR image.

    \item SPGSR combines visually guided residual prediction with
    full-resolution refinement and learns directly from paired images
    without SR teacher distillation.

    \item With 31.8M generator parameters, SPGSR improves PSNR, SSIM,
    LPIPS, and DISTS over VOSR-0.5B on all three benchmarks and reduces
    GPU-pipeline latency from 50.47 to 18.05\,ms for $512\times512$
    outputs on an RTX PRO 6000 under FP16 inference.
\end{itemize}
\section{Related Work}
\label{sec:related_work}

\subsection{T2I-Based Super-Resolution}

T2I-based SR methods adapt pretrained text-to-image diffusion models to reconstruct HR images from LR inputs.
StableSR~\cite{wang2024exploiting} trains a time-aware encoder and feature-modulation layers around a frozen Stable Diffusion backbone, and introduces controllable feature wrapping to balance perceptual quality and fidelity.
DiffBIR~\cite{lin2024diffbir} separates degradation removal from information regeneration, using IRControlNet to condition the generation stage on an initial restoration.
SeeSR~\cite{wu2024seesr} addresses semantic ambiguity with degradation-aware semantic prompts.

One-step methods reduce iterative sampling while retaining pretrained generative priors.
OSEDiff~\cite{wu2024one} initializes generation from the encoded LR image and uses latent-space variational score distillation as a distribution regularizer.
S3Diff~\cite{zhang2024s3diff} employs degradation-guided low-rank adaptation, while TSD-SR~\cite{dong2025tsdsr} introduces target score distillation and distribution-aware sampling.
PiSA-SR~\cite{sun2025pisa} uses two LoRA modules to separate pixel-level regression from semantic enhancement, enabling adjustable fidelity and perceptual quality at inference without retraining.
FiDeSR~\cite{kim2026fidesr} combines detail-aware loss weighting, adaptive enhancement of low and high frequencies, and latent residual refinement to correct errors in the predicted residual.

Beyond reducing sampling steps, AdcSR~\cite{chen2025adcsr} compresses OSEDiff through module removal and channel pruning, using adversarial distillation to preserve its generative capabilities.
TinySR~\cite{dong2025tinysr} compresses TSD-SR through transformer depth pruning, VAE compression, and pre-caching of input-independent modulation parameters.
PocketSR~\cite{sun2025pocketsr} builds on SD-Turbo and achieves efficient restoration through a lightweight encoder-decoder, online annealing pruning, and multi-layer feature distillation.

\subsection{Vision-Only Super-Resolution}

Generative SR models can learn restoration directly from paired LR and HR images without relying on pretrained T2I generators.
SRGAN~\cite{ledig2017photo} and ESRGAN~\cite{wang2018esrgan} use content and adversarial losses to learn perceptually realistic reconstructions.
Real-ESRGAN~\cite{wang2021realesrgan} extends this approach to blind SR with training pairs synthesized using high-order degradations.
SR3~\cite{saharia2022image} and SRDiff~\cite{li2022srdiff} use conditional diffusion to recover HR images through iterative denoising.
ResShift~\cite{yue2023resshift} improves sampling efficiency through residual shifting.
SinSR~\cite{wang2024sinsr} and RSD~\cite{selikhanovych2026rsd} distill ResShift-based models into one-step restorers.

Pretrained visual representations provide semantic guidance for SR.
DINOv2~\cite{oquab2023dinov2} learns visual representations through self-supervised training without text supervision.
DINO-IR~\cite{lin2024dinoir} uses DINO-based contrastive supervision for image restoration, whereas our feature objective aligns paired SR--HR features through cosine similarity without explicit negative samples.
Projected GANs~\cite{sauer2021projected} establish pretrained-feature adversarial discrimination.
Our DINO-based discriminator follows this broader principle rather than adopting their discriminator architecture.
VOSR~\cite{wu2026vosr} combines visual semantic conditioning with LR-derived structural conditioning in VAE latent space.
It introduces restoration-oriented guidance that preserves weakened LR structural cues in the partially conditioned branch rather than using a fully unconditional branch.
VOSR trains its generative backbone from scratch and subsequently distills the multi-step model into a one-step restorer, while investigating model scaling, restoration quality, and training efficiency.
SPGSR follows this vision-only direction with a compact pixel-space generator, learning one-step restoration under a limited generator training budget without SR teacher distillation.

\subsection{Pixel-Space Generation}

DDPM~\cite{ho2020denoising} models image generation directly in pixel space.
Latent diffusion~\cite{rombach2022high} reduces generative modeling costs by operating in the compressed representations of a pretrained autoencoder, and DiT~\cite{peebles2023scalable} introduces a scalable transformer backbone operating on latent patches.
Recent works revisit pixel-space generation with transformer-based architectures and perceptual supervision.
JiT~\cite{li2025jit} demonstrates that transformers operating on large pixel patches can generate images by predicting clean data rather than noise or velocity.
DiP~\cite{chen2026dip} combines a transformer for global structure with a lightweight patch detailer for local synthesis.
For image synthesis rather than super-resolution, PixelGen~\cite{ma2026pixelgen} augments clean-image prediction with LPIPS and DINO-based perceptual supervision, applied at lower-noise timesteps.
These studies motivate pixel-space architectures and perceptual objectives, but their image synthesis settings do not by themselves establish faithful restoration of a degraded observation. We study how a compact pixel-space generator trained from scratch can achieve high-quality image super-resolution using pretrained visual representations for conditioning and lightweight local residual refinement.

\section{Method}
\label{sec:method}

\begin{figure*}[t]
    \centering
    \includegraphics[width=\textwidth]{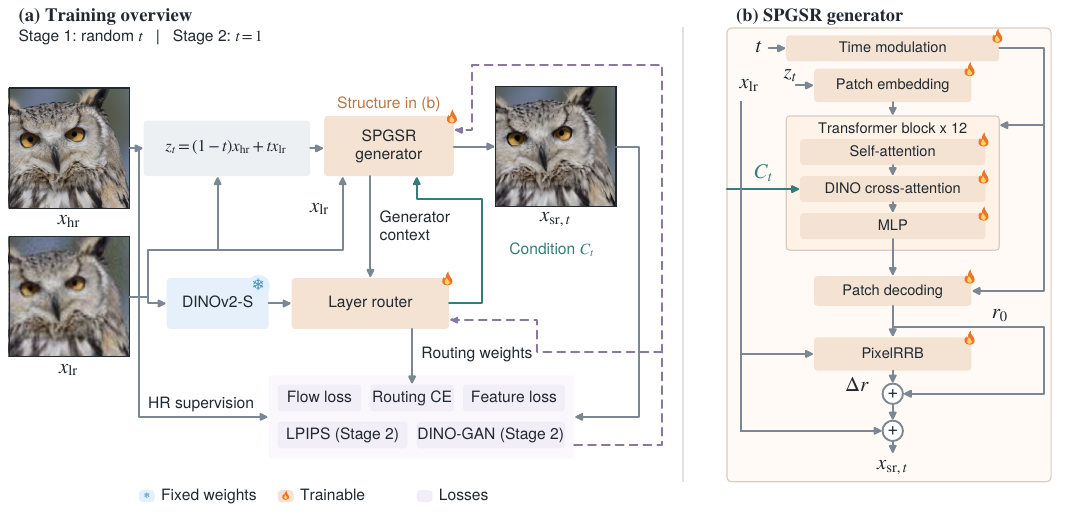}
    \caption{\textbf{Overview of SPGSR.}
    (a) Two-stage training with frozen DINO features and trainable layer routing.
    (b) Generator architecture with DINO-conditioned Transformer blocks and PixelRRB refinement.
    Orange modules are trainable, blue encoders are frozen, and dashed arrows denote optimization.
    Thumbnails show bicubic-upsampled LR, HR, and raw one-step SR.}
    \label{fig:framework}
\end{figure*}

The LR image already contains information about the scene structure. SPGSR uses its upsampled version as an initial estimate and adds a learned RGB correction to remove artifacts and recover missing detail. Predicting that correction calls for both visual context to interpret degraded structures and local pixel information to refine their appearance. Inspired by PixRestore~\cite{sun2026pixrestore}, we use a pixel-space transformer to predict a patchwise RGB residual with guidance from frozen DINOv2 features~\cite{oquab2023dinov2}. A lightweight refinement block then uses neighboring pixels in the upsampled image and the predicted residual to refine the correction across patch boundaries (Fig.~\ref{fig:framework}).
\subsection{LR-Anchored Pixel-Space Generation}
\label{sec:lr_anchor}
Given a LR image $y$ and its paired HR target $x_{\mathrm{hr}}$, we denote the target-resolution LR condition by $x_{\mathrm{lr}}$ and normalize both $x_{\mathrm{lr}}$ and $x_{\mathrm{hr}}$ to $[-1,1]$.
At inference, $x_{\mathrm{lr}}=\mathcal{U}_4(y)$ is bicubic-upsampled.
The training resampling recipe is in Sec.~\ref{sec:experimental_settings}.
Given a restoration state $z_t$ at time $t$, we always add the predicted residual back to the same LR image:
\begin{equation}
    x_{\mathrm{sr},t}=G_\theta(z_t,t;x_{\mathrm{lr}})=x_{\mathrm{lr}}+r_\theta(z_t,t;x_{\mathrm{lr}}),
    \label{eq:sr_prediction}
\end{equation}
$r_\theta$ includes transformer prediction and local refinement.

\paragraph{Restoration path.}
At inference, the generator must predict the full correction from the LR image alone. During training, the paired HR image lets us vary how much target detail is visible in the input while keeping the target residual $x_{\mathrm{hr}}-x_{\mathrm{lr}}$ fixed. We construct these training inputs by linearly interpolating between the paired HR and LR images~\cite{lipman2023flow,liu2023rectifiedflow},
\begin{equation}
    z_t=(1-t)x_{\mathrm{hr}}+t x_{\mathrm{lr}},\qquad t\in[0,1].
    \label{eq:sr_path}
\end{equation}
no independent Gaussian noise is added.
We train the RGB prediction with a velocity-space objective derived from this path~\cite{li2025jit,sun2026pixrestore}.
For our LR-anchored path, we match the stabilized velocity surrogate $(z_t-x_{\mathrm{sr},t})/\bar t$ to its target $(z_t-x_{\mathrm{hr}})/\bar t$, where $\bar t=\max(t,0.05)$, yielding
\begin{equation}
    \mathcal{L}_{\mathrm{flow}}
    =\mathbb{E}_{x_{\mathrm{hr}},x_{\mathrm{lr}},t}\left[
      \frac{\|x_{\mathrm{sr},t}-x_{\mathrm{hr}}\|_{\mathrm{avg}}^2}{\bar t^2}
    \right],
    \label{eq:sr_flow}
\end{equation}
where $\|\cdot\|_{\mathrm{avg}}^2$ denotes the mean squared value over image elements.
The denominator is clamped near $t=0$ for numerical stability.
At inference, we obtain $x_{\mathrm{sr}}=x_{\mathrm{sr},1}=G_\theta(x_{\mathrm{lr}},1;x_{\mathrm{lr}})$ with one visual-encoder pass and one generator evaluation, without iterative integration.

\paragraph{Generator backbone.}
To model image-wide context with fewer tokens than per-pixel processing, we patchify the three-channel LR-anchored state $z_t$ into non-overlapping patches and process the tokens with a transformer.
Time embeddings modulate the blocks, while cross-attention injects the LR-derived visual condition defined in Sec.~\ref{sec:visual_guidance}.
The output tokens are projected and unpatchified into the full-resolution base RGB residual $r_0$, which is refined by PixelRRB (Sec.~\ref{sec:pixel_refinement}).

\subsection{Visual Conditioning and Feature Supervision}
\label{sec:visual_guidance}
Based on comparisons with single-layer conditioning, uniform fusion, and unsupervised routing (Sec.~\ref{sec:ablations}), we adopt LR--HR-agreement-guided layer routing inspired by PixRestore~\cite{sun2026pixrestore}.
During training, we compare features from paired LR and HR images
at each layer of a frozen DINOv2-S encoder~\cite{oquab2023dinov2}.
Layers with high LR--HR agreement provide stable visual cues for
conditioning the generator. Layers with lower agreement are more
sensitive to degradation, so we give them greater weight when
matching the reconstructed image to its HR target in feature space.
\paragraph{Layer-routed conditioning.}
To obtain these fusion weights without an HR reference at inference, we train an input-dependent router using targets derived from layer-wise LR--HR agreement during training.

Let $E_l(\cdot)$ denote encoder features from layer $l$ and $F_l(\cdot)$ their RMS-normalized versions.
Each layer's score $s_l$ averages rescaled cosine and normalized-error similarities between its normalized LR and HR features~\cite{sun2026pixrestore}.
With $\tau=1$, we form detached routing targets:
\begin{equation}
    \mathbf{q}=\operatorname{softmax}(\mathbf{s}/\tau).
    \label{eq:routing_target}
\end{equation}

A shared trainable projection $P$ maps the normalized features to conditioning features.
The router context $h_t$ combines the restoration state, time embedding, and LR-derived features.
The router $g_\theta$ predicts fusion weights over the $L$ selected layers:
\begin{equation}
    \mathbf{p}=\operatorname{softmax}(g_\theta(h_t)),
    \qquad
    C_t=\sum_{l=1}^{L}p_l P(F_l(x_{\mathrm{lr}})).
    \label{eq:visual_fusion}
\end{equation}
$C_t$ conditions each transformer block through cross-attention, with $\mathbf{p}$ shared across blocks and spatial positions.
We supervise $\mathbf{p}$ with $\mathcal{L}_{\mathrm{route}}=-\mathbb{E}\sum_l q_l\log p_l$.
At inference, no HR features or $\mathbf{q}$ are used.

\paragraph{Layer-weighted feature supervision.}
To focus on representations more affected by degradation, feature supervision favors layers with larger LR--HR discrepancies:
$w_l=\operatorname{softmax}_l((\mathbf{1}-\mathbf{s})/\tau)$.
We align SR and HR features with a weighted cosine loss:
\begin{equation}
    \mathcal{L}_{\mathrm{feat}}
    =\mathbb{E}\sum_l \frac{w_l}{N}\sum_n
    \left[1-\cos\!\left(E_l(x_{\mathrm{sr},t})_n,E_l(x_{\mathrm{hr}})_n\right)\right].
    \label{eq:feature_supervision}
\end{equation}
Here $N$ is the number of encoder patch tokens.
DINO remains frozen while this loss updates the generator.

\subsection{Pixel Residual Refinement}
\label{sec:pixel_refinement}

The transformer exchanges context between patch tokens, but its output projection predicts pixels patch by patch.
We add a lightweight Pixel Residual Refinement Block (PixelRRB) so that neighboring output pixels on either side of a patch boundary can inform the final correction.
This local residual update builds on residual learning~\cite{he2016deep} and its use in SR~\cite{zhang2018rcan,wang2018esrgan}.
PixelRRB takes both the LR condition and the unpatchified residual $r_0$, making its correction depend on observed structure and the current residual prediction:
\begin{equation}
    \Delta r=R_\psi([x_{\mathrm{lr}},r_0]),\qquad
    r_\theta=r_0+\Delta r.
    \label{eq:pixel_refinement}
\end{equation}
Here $R_\psi$ denotes PixelRRB and $[\cdot,\cdot]$ denotes channel concatenation.
Its four 64-channel convolutional residual blocks operate at the full output resolution.
We zero-initialize the output head so refinement initially leaves $r_0$ unchanged, and train the module jointly with the transformer in both stages.
Related residual correction is used in FiDeSR~\cite{kim2026fidesr}, whose RRDB-based module refines a predicted latent residual before VAE decoding.
PixelRRB refines the RGB residual within the same generator evaluation.

\subsection{Two-Stage Training}
\label{sec:training_inference}

\paragraph{Stage 1: restoration training.}
We first learn to reconstruct HR images from states containing different amounts of target detail.
We sample $t\sim\mathcal{U}(0,1)$ along Eq.~\eqref{eq:sr_path} and combine image reconstruction with routing and feature supervision:
\begin{equation}
    \mathcal{L}_{\mathrm{base}}
    =\mathcal{L}_{\mathrm{flow}}
      +\lambda_{\mathrm{route}}\mathcal{L}_{\mathrm{route}}
      +\lambda_{\mathrm{feat}}\mathcal{L}_{\mathrm{feat}}.
    \label{eq:base_objective}
\end{equation}
The generator is randomly initialized, and the visual encoder remains frozen.

\paragraph{Stage 2: fixed-time perceptual refinement.}
Since inference starts from the LR input, we finetune the generator using the LR image as the input in Stage 2.
We initialize the generator from its Stage 1 EMA weights and fix $t=1$.
To encourage perceptually realistic detail while retaining paired reconstruction supervision, we keep the base objectives and add VGG-based LPIPS~\cite{zhang2018perceptual} and DINO-feature adversarial loss:
\begin{equation}
    \mathcal{L}_{G}
    =\mathcal{L}_{\mathrm{base}}
      +\lambda_{\mathrm{LPIPS}}\mathcal{L}_{\mathrm{LPIPS}}
      +\lambda_{\mathrm{adv}}\mathcal{L}_{\mathrm{adv}}.
    \label{eq:stage2_objective}
\end{equation}
LPIPS penalizes perceptual differences between SR and HR images.
We use a token-wise MLP discriminator head at each selected frozen DINO layer to distinguish reconstructed images from real HR images in feature space~\cite{sun2026pixrestore}.
Adversarial supervision uses binary cross-entropy averaged over tokens within each layer and summed across layers.
The generator loss $\mathcal{L}_{\mathrm{adv}}$ encourages SR predictions to be classified as real.
The discriminator minimizes the summed losses for real HR images and detached SR predictions labeled as fake, weighted by $\lambda_{\mathrm{adv}}$.
Both updates use a smoothed real target.
We alternate generator and discriminator updates, freezing the discriminator heads during generator updates and keeping the DINO encoder frozen throughout.

\paragraph{Condition weakening during training.}
To reduce dependence on full-strength conditioning, we train the generator to reconstruct the same HR target when DINO guidance is absent and LR inputs are attenuated.
We retain these weakened LR cues as a spatial reference, adapting VOSR's condition-weakening strategy~\cite{wu2026vosr}.
In Stage 2, with probability $0.1$ per sample, we disable DINO cross-attention and scale the normalized LR image by $0.3$ in the input state, LR descriptor, PixelRRB input, and output skip connection.
Unlike VOSR, we also weaken the input state itself.
For these samples, we omit routing supervision but retain all other training objectives.
Feature-loss weights are computed from the original LR--HR pair.
At inference, we enable DINO cross-attention and use the normalized LR image without attenuation.

\section{Experiments}
\label{sec:experiments}

\begin{table*}[t]
    \centering
    \caption{Quantitative SR comparison on three benchmarks.
    Except for PocketSR, all methods are evaluated locally on the same test pairs with a common metric implementation.
    $^{\dagger}$PocketSR results are taken from its paper and excluded from ranking.
    -- denotes an unreported metric.
    ``-Ns'' denotes $N$ restoration steps.
    \textbf{VO} denotes methods without pretrained T2I generators.
    \textbf{T2I} denotes methods adapted or distilled from pretrained text-to-image generators.
    The top three distinct displayed values per dataset and metric are marked in bold red, bold blue, and bold underline, respectively.}
    \label{tab:sr_quantitative}
    \small
    \setlength{\tabcolsep}{3pt}
    \begin{tabular}{llcrrrrrrrrr}
        \toprule
        Dataset & Method & Type & PSNR$\uparrow$ & SSIM$\uparrow$ & LPIPS$\downarrow$ & DISTS$\downarrow$ & CLIPIQA$\uparrow$ & NIQE$\downarrow$ & MUSIQ$\uparrow$ & MANIQA$\uparrow$ & FID$\downarrow$ \\
        \midrule
         & SeeSR-50s~\cite{wu2024seesr} &  & 28.24 & 0.7695 & 0.3211 & 0.2324 & 0.6807 & 6.7071 & 64.4669 & 0.5967 & 147.50 \\
         & DiffBIR-50s~\cite{lin2024diffbir} &  & 25.73 & 0.6111 & 0.4794 & 0.2881 & 0.7053 & 6.4530 & \textcolor{red}{\textbf{67.6515}} & \textcolor{blue}{\textbf{0.6252}} & 168.33 \\
         & OSEDiff-1s~\cite{wu2024one} &  & 27.91 & \textcolor{blue}{\textbf{0.7833}} & 0.2968 & \underline{\textbf{0.2164}} & 0.6972 & 6.4494 & 64.6973 & 0.5903 & 135.60 \\
         & PiSA-SR-1s~\cite{sun2025pisa} &  & \underline{\textbf{28.32}} & \underline{\textbf{0.7804}} & \underline{\textbf{0.2960}} & 0.2169 & 0.6975 & 6.1739 & 66.1081 & 0.6161 & \textcolor{blue}{\textbf{130.47}} \\
         & FiDeSR-1s~\cite{kim2026fidesr} &  & \textcolor{red}{\textbf{29.03}} & \textcolor{red}{\textbf{0.7941}} & \textcolor{red}{\textbf{0.2811}} & \textcolor{red}{\textbf{0.2101}} & 0.6919 & 6.2339 & 65.3542 & 0.6184 & \textcolor{red}{\textbf{127.50}} \\
         & AdcSR-1s~\cite{chen2025adcsr} &  & 28.10 & 0.7726 & 0.3046 & 0.2200 & 0.7049 & 6.4471 & 66.2649 & 0.5916 & \underline{\textbf{134.02}} \\
         & CODSR-1s~\cite{chen2026codsr} &  & 28.19 & 0.7765 & \textcolor{blue}{\textbf{0.2916}} & \textcolor{blue}{\textbf{0.2109}} & \textcolor{red}{\textbf{0.7153}} & 5.9646 & \textcolor{blue}{\textbf{67.0258}} & \textcolor{red}{\textbf{0.6287}} & 136.65 \\
         & TinySR-1s~\cite{dong2025tinysr} &  & 27.48 & 0.7459 & 0.3116 & 0.2205 & \textcolor{blue}{\textbf{0.7090}} & \textcolor{red}{\textbf{5.6727}} & 65.3588 & 0.5779 & 146.82 \\
         & PocketSR-1s$^{\dagger}$~\cite{sun2025pocketsr} & \multirow{-9}{*}{T2I} & 28.05 & 0.7675 & 0.2962 & 0.2139 & -- & 5.809 & 63.85 & -- & -- \\
        \rowcolor{gray!15} \cellcolor{white} & SinSR-1s~\cite{wang2024sinsr} &  & 28.11 & 0.7509 & 0.3560 & 0.2481 & 0.6613 & 6.7455 & 57.3725 & 0.5057 & 175.36 \\
        \rowcolor{gray!15} \cellcolor{white} & VOSR-0.5B-1s~\cite{wu2026vosr} &  & 27.31 & 0.7343 & 0.3543 & 0.2418 & 0.6131 & \underline{\textbf{5.8324}} & 66.0535 & 0.6108 & 169.97 \\
        \rowcolor{gray!15} \cellcolor{white} & VOSR-1.4B-1s~\cite{wu2026vosr} &  & 27.08 & 0.7152 & 0.3458 & 0.2381 & 0.6221 & 6.0249 & \underline{\textbf{66.3982}} & 0.6058 & 163.54 \\
        \rowcolor{gray!15} \cellcolor{white}\multirow{-13}{*}{\textbf{DRealSR}} & \textbf{SPGSR-1s (Ours)} & \multirow{-4}{*}{VO} & \textcolor{blue}{\textbf{28.53}} & 0.7729 & 0.3124 & 0.2193 & \underline{\textbf{0.7079}} & \textcolor{blue}{\textbf{5.7070}} & 66.1135 & \underline{\textbf{0.6237}} & 138.36 \\
        \midrule
         & SeeSR-50s~\cite{wu2024seesr} &  & 25.31 & 0.7247 & 0.2968 & 0.2222 & 0.6723 & 5.3887 & 69.5645 & 0.6460 & 124.31 \\
         & DiffBIR-50s~\cite{lin2024diffbir} &  & 24.83 & 0.6532 & 0.3595 & 0.2398 & \textcolor{red}{\textbf{0.7097}} & 5.7591 & 68.9581 & 0.6495 & 128.97 \\
         & OSEDiff-1s~\cite{wu2024one} &  & 25.15 & 0.7338 & 0.2921 & 0.2129 & 0.6688 & 5.6438 & 69.0833 & 0.6337 & 123.52 \\
         & PiSA-SR-1s~\cite{sun2025pisa} &  & \underline{\textbf{25.50}} & \underline{\textbf{0.7417}} & \underline{\textbf{0.2672}} & 0.2044 & 0.6698 & 5.5021 & \underline{\textbf{70.1471}} & 0.6551 & 124.13 \\
         & FiDeSR-1s~\cite{kim2026fidesr} &  & \textcolor{red}{\textbf{26.16}} & \textcolor{blue}{\textbf{0.7484}} & \textcolor{red}{\textbf{0.2611}} & \textcolor{red}{\textbf{0.1961}} & 0.6821 & 5.3302 & 69.5777 & \underline{\textbf{0.6631}} & \textcolor{red}{\textbf{109.49}} \\
         & AdcSR-1s~\cite{chen2025adcsr} &  & 25.47 & 0.7301 & 0.2885 & 0.2129 & 0.6731 & 5.3480 & 69.8999 & 0.6353 & 118.43 \\
         & CODSR-1s~\cite{chen2026codsr} &  & 25.36 & 0.7276 & 0.2747 & \underline{\textbf{0.2006}} & \underline{\textbf{0.6894}} & 5.2829 & \textcolor{blue}{\textbf{70.4868}} & \textcolor{red}{\textbf{0.6735}} & 118.07 \\
         & TinySR-1s~\cite{dong2025tinysr} &  & 24.79 & 0.7171 & 0.2806 & 0.2123 & \textcolor{blue}{\textbf{0.7038}} & \textcolor{red}{\textbf{4.7405}} & 69.7787 & 0.6212 & 118.18 \\
         & PocketSR-1s$^{\dagger}$~\cite{sun2025pocketsr} & \multirow{-9}{*}{T2I} & 25.47 & 0.7330 & 0.2713 & 0.2094 & -- & 5.067 & 67.07 & -- & -- \\
        \rowcolor{gray!15} \cellcolor{white} & SinSR-1s~\cite{wang2024sinsr} &  & \textcolor{blue}{\textbf{26.12}} & 0.7361 & 0.3083 & 0.2326 & 0.6282 & 6.1203 & 61.2424 & 0.5445 & 135.71 \\
        \rowcolor{gray!15} \cellcolor{white} & VOSR-0.5B-1s~\cite{wu2026vosr} &  & 24.85 & 0.7096 & 0.2904 & 0.2107 & 0.6015 & \underline{\textbf{5.1429}} & 69.9140 & 0.6540 & 122.47 \\
        \rowcolor{gray!15} \cellcolor{white} & VOSR-1.4B-1s~\cite{wu2026vosr} &  & 24.55 & 0.7023 & 0.2869 & 0.2127 & 0.6032 & 5.4815 & \textcolor{red}{\textbf{71.0567}} & 0.6524 & \textcolor{blue}{\textbf{113.09}} \\
        \rowcolor{gray!15} \cellcolor{white}\multirow{-13}{*}{\textbf{RealSR}} & \textbf{SPGSR-1s (Ours)} & \multirow{-4}{*}{VO} & \textcolor{blue}{\textbf{26.12}} & \textcolor{red}{\textbf{0.7507}} & \textcolor{blue}{\textbf{0.2638}} & \textcolor{blue}{\textbf{0.2001}} & 0.6765 & \textcolor{blue}{\textbf{5.1008}} & 70.0453 & \textcolor{blue}{\textbf{0.6639}} & \underline{\textbf{113.68}} \\
        \midrule
         & SeeSR-50s~\cite{wu2024seesr} &  & 23.68 & 0.6046 & 0.3198 & 0.1963 & 0.6941 & 4.8230 & 68.6412 & 0.6218 & 25.96 \\
         & DiffBIR-50s~\cite{lin2024diffbir} &  & 23.16 & 0.5451 & 0.3654 & 0.2204 & \textcolor{red}{\textbf{0.7293}} & 5.0146 & \textcolor{blue}{\textbf{69.9016}} & \textcolor{blue}{\textbf{0.6438}} & 32.37 \\
         & OSEDiff-1s~\cite{wu2024one} &  & 23.72 & \underline{\textbf{0.6109}} & 0.2942 & 0.1975 & 0.6682 & 4.7092 & 67.9714 & 0.6132 & 26.34 \\
         & PiSA-SR-1s~\cite{sun2025pisa} &  & 23.87 & 0.6058 & 0.2823 & 0.1934 & 0.6928 & 4.5570 & 69.6838 & 0.6375 & 25.09 \\
         & FiDeSR-1s~\cite{kim2026fidesr} &  & \textcolor{red}{\textbf{24.43}} & \textcolor{red}{\textbf{0.6276}} & \textcolor{red}{\textbf{0.2681}} & \textcolor{blue}{\textbf{0.1844}} & 0.6761 & 4.6575 & 68.5071 & 0.6331 & \textcolor{blue}{\textbf{23.26}} \\
         & AdcSR-1s~\cite{chen2025adcsr} &  & 23.74 & 0.6017 & 0.2853 & 0.1899 & 0.6764 & \textcolor{blue}{\textbf{4.3580}} & 67.9975 & 0.6073 & 25.52 \\
         & CODSR-1s~\cite{chen2026codsr} &  & 23.79 & 0.6065 & 0.2878 & 0.1929 & 0.6936 & 4.6529 & 69.6055 & \underline{\textbf{0.6412}} & 24.56 \\
         & TinySR-1s~\cite{dong2025tinysr} &  & 22.76 & 0.5725 & \underline{\textbf{0.2793}} & \underline{\textbf{0.1883}} & \textcolor{blue}{\textbf{0.7201}} & \textcolor{red}{\textbf{4.1509}} & \underline{\textbf{69.8970}} & 0.6050 & \underline{\textbf{24.45}} \\
         & PocketSR-1s$^{\dagger}$~\cite{sun2025pocketsr} & \multirow{-9}{*}{T2I} & 23.85 & 0.6015 & 0.2801 & 0.1830 & -- & 4.415 & 66.38 & -- & 25.25 \\
        \rowcolor{gray!15} \cellcolor{white} & SinSR-1s~\cite{wang2024sinsr} &  & \underline{\textbf{24.28}} & 0.6000 & 0.3233 & 0.2054 & 0.6560 & 5.8108 & 63.3340 & 0.5417 & 37.18 \\
        \rowcolor{gray!15} \cellcolor{white} & VOSR-0.5B-1s~\cite{wu2026vosr} &  & 23.42 & 0.5837 & 0.3209 & 0.1957 & 0.6045 & \underline{\textbf{4.3606}} & 69.3397 & 0.6051 & 27.04 \\
        \rowcolor{gray!15} \cellcolor{white} & VOSR-1.4B-1s~\cite{wu2026vosr} &  & 23.27 & 0.5824 & 0.3215 & 0.1942 & 0.5862 & 4.9528 & \textcolor{red}{\textbf{70.2147}} & 0.6120 & 27.02 \\
        \rowcolor{gray!15} \cellcolor{white}\multirow{-13}{*}{\textbf{DIV2K}} & \textbf{SPGSR-1s (Ours)} & \multirow{-4}{*}{VO} & \textcolor{blue}{\textbf{24.38}} & \textcolor{blue}{\textbf{0.6140}} & \textcolor{blue}{\textbf{0.2752}} & \textcolor{red}{\textbf{0.1795}} & \underline{\textbf{0.7051}} & 4.5717 & 69.5395 & \textcolor{red}{\textbf{0.6444}} & \textcolor{red}{\textbf{21.87}} \\
        \bottomrule
    \end{tabular}
\end{table*}

\begin{figure*}[t]
    \centering
    \includegraphics[width=\textwidth]{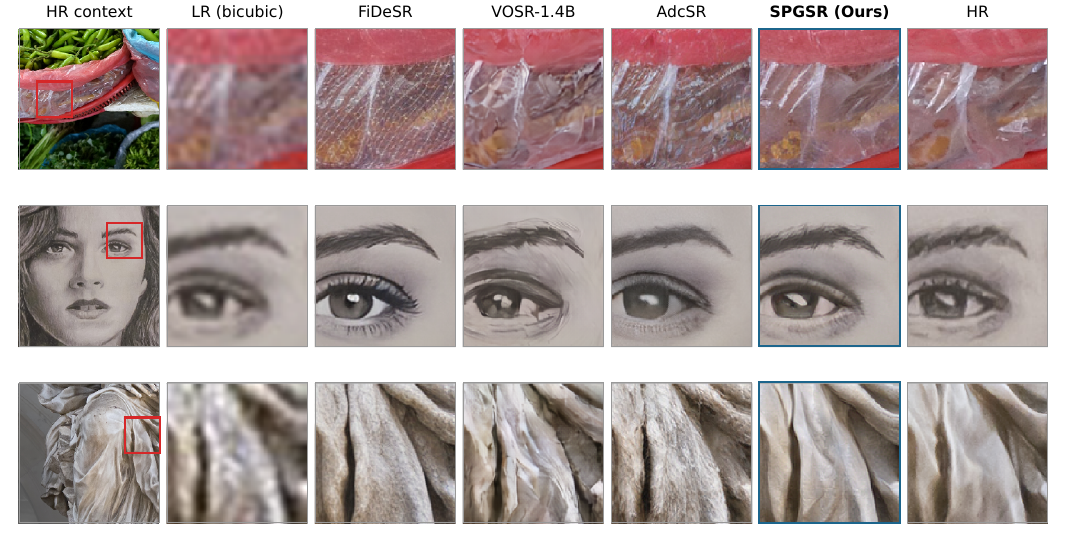}
    \caption{Selected visual comparisons, from top to bottom: transparent plastic folds (DIV2K), a pencil-drawing eye (RealSR), and statue drapery (DIV2K).
    Red boxes mark the shared $128\times128$ HR-coordinate crops, shown at the same scale.}
    \label{fig:visual_comparison}
\end{figure*}

\subsection{Experimental Settings}
\label{sec:experimental_settings}

\paragraph{Training settings.}
We train SPGSR for $\times4$ SR on LSDIR~\cite{li2023lsdir}, DIV2K~\cite{agustsson2017div2k}, Flickr2K~\cite{lim2017edsr}, and the first 10K FFHQ images~\cite{karras2019style}.
Following PiSA-SR's emphasis on high-quality training data~\cite{sun2025pisa}, each training draw samples from a fixed high-quality LSDIR subset with probability $0.9$ or from the full pool of these datasets with probability $0.1$.
We generate training pairs using Real-ESRGAN degradation~\cite{wang2021realesrgan}, random $512\times512$ HR crops, and horizontal flips.

The generator uses $8\times8$ patches and 12 transformer blocks of width 384 with six attention heads.
DINOv2-S remains frozen for both training stages.
In Stage 1, we train the randomly initialized generator for 55k optimizer updates with a global batch size of 8.
Stage 2 starts from the Stage 1 EMA checkpoint and trains for 50k updates with a global batch size of 16.

Both stages use AdamW~\cite{loshchilov2019adamw} with a constant learning rate of 1e-4 on two NVIDIA RTX PRO 6000 Blackwell GPUs.
We set $\lambda_{\mathrm{route}}=\lambda_{\mathrm{feat}}=0.5$ in both stages and $\lambda_{\mathrm{LPIPS}}=\lambda_{\mathrm{adv}}=0.1$ in Stage 2.
We count generator training pixels as optimizer updates $\times$ global batch size $\times$ HR crop area, giving approximately $1.15\times10^{11}$ for Stage 1 and $2.10\times10^{11}$ for Stage 2.
Condition-weakening settings are described in Sec.~\ref{sec:training_inference}.

\paragraph{Compared methods.}
We compare SPGSR with vision-only one-step methods SinSR~\cite{wang2024sinsr}, VOSR-0.5B, and VOSR-1.4B~\cite{wu2026vosr}.
T2I-based one-step baselines include OSEDiff~\cite{wu2024one}, PiSA-SR~\cite{sun2025pisa}, FiDeSR~\cite{kim2026fidesr}, and CODSR~\cite{chen2026codsr}.
We also compare with compact generative SR methods AdcSR~\cite{chen2025adcsr}, TinySR~\cite{dong2025tinysr}, and PocketSR~\cite{sun2025pocketsr}.
SeeSR~\cite{wu2024seesr} and DiffBIR~\cite{lin2024diffbir}, both using 50 restoration steps, serve as multi-step baselines.
Except for PocketSR, these methods are evaluated locally on the same test pairs with a common metric implementation.
We use their official repositories, released checkpoints, and inference pipelines.
We use PocketSR's paper-reported results.

\paragraph{Test datasets and evaluation metrics.}
We evaluate on 3,000 synthetic DIV2K validation pairs, 100 RealSR pairs~\cite{cai2019realsr}, and 93 DRealSR pairs~\cite{wei2020drealsr}.
All pairs have $128\times128$ LR inputs and $512\times512$ HR targets.
We use the released pairs without regenerating degradations or modifying HR targets.
We report PSNR/SSIM~\cite{wang2004ssim} for fidelity, LPIPS~\cite{zhang2018perceptual}/DISTS~\cite{ding2020dists} for perceptual similarity, and FID~\cite{heusel2017fid} for distributional similarity.
NIQE~\cite{mittal2013niqe}, MUSIQ~\cite{ke2021musiq}, MANIQA~\cite{yang2022maniqa}, and CLIPIQA~\cite{wang2023clipiqa} provide no-reference quality assessments.
PSNR and SSIM are evaluated on the Y channel without border cropping.

\subsection{Comparison with Existing Methods}
\label{sec:comparisons}

\paragraph{Quantitative results.}
Among the evaluated vision-only methods, SPGSR achieves the highest SSIM, CLIPIQA and MANIQA and the lowest LPIPS and DISTS on all three datasets (Table~\ref{tab:sr_quantitative}).
It also leads this group in PSNR on DIV2K and DRealSR and ties with SinSR in reported PSNR on RealSR.
VOSR-1.4B has higher MUSIQ on all three datasets and lower FID on RealSR, while VOSR-0.5B has lower NIQE on DIV2K.
SPGSR uses only 31.8M generator parameters, corresponding to 6.5\% and 2.3\% of the generator parameters of the VOSR-0.5B and VOSR-1.4B \mbox{models, respectively}.

Across all locally evaluated methods in Table~\ref{tab:sr_quantitative}, SPGSR achieves the lowest DISTS (0.1795) and FID (21.87) and the highest MANIQA (0.6444) on DIV2K, and the highest SSIM on RealSR.
On DRealSR, it ranks second in PSNR and NIQE and third in CLIPIQA and MANIQA.

\begin{table}[t]
\centering
\caption{One-step FP16 inference on RTX PRO 6000 Blackwell, batch 1, $128\!\rightarrow\!512$. Gen.\ and All denote generator and total loaded parameters. GPU reports full-pipeline latency, and Mem.\ reports peak allocated memory. Best and second-best values are bold and underlined, respectively.}
\label{tab:sr_efficiency}
\footnotesize
\setlength{\tabcolsep}{2pt}
\begin{tabular}{lrrrr}
\toprule
Method & Gen.\ (M) & All (M) & GPU (ms)$\downarrow$ & Mem.\ (GiB)$\downarrow$ \\
\midrule
\multicolumn{5}{l}{\textit{FP16 weights, native autocast policy}} \\
SinSR & \underline{118.59} & \underline{173.92} & 70.25 & 1.732 \\
OSEDiff & 869.97 & 1765.30 & 68.56 & 3.984 \\
PiSA-SR & 865.91 & 1289.95 & 51.59 & 4.662 \\
FiDeSR & 866.63 & 1290.67 & 53.27 & 3.038 \\
TinySR & 361.44 & 362.83 & \textbf{10.11} & \underline{0.849} \\
VOSR-0.5B & 486.94 & 670.18 & 50.47 & 1.672 \\
VOSR-1.4B & 1396.70 & 1745.67 & 60.46 & 3.628 \\
\textbf{SPGSR} & \textbf{31.80} & \textbf{53.86} & \underline{18.05} & \textbf{0.293} \\
\bottomrule
\end{tabular}
\end{table}

\paragraph{Inference efficiency.}
Table~\ref{tab:sr_efficiency} compares full-pipeline one-step inference costs, including the frozen encoder and required decoding modules.
With 53.86M loaded parameters (31.80M in the generator), SPGSR uses 92.0\% and 96.9\% fewer loaded parameters than VOSR-0.5B and VOSR-1.4B, respectively.
Under FP16 inference, its 18.05\,ms GPU-pipeline latency gives $2.80\times$ and $3.35\times$ speedups over VOSR-0.5B and VOSR-1.4B, respectively.

\paragraph{Comparison with compact generative SR.}
SPGSR achieves higher PSNR/SSIM and lower DISTS than AdcSR and TinySR on all three benchmarks (Table~\ref{tab:sr_quantitative}).
Compared with TinySR, it also achieves higher MANIQA and lower FID, with 91.2\% fewer generator parameters and 65.5\% lower peak GPU memory (Table~\ref{tab:sr_efficiency}).
SPGSR also uses 88.2\% fewer total inference parameters than AdcSR and 63.2\% fewer than PocketSR's paper-reported 146.2M model~\cite{sun2025pocketsr}.
Relative to PocketSR's paper-reported results, SPGSR achieves higher PSNR/SSIM and MUSIQ on all three benchmarks, as well as lower LPIPS/DISTS on DIV2K and RealSR.

\paragraph{Qualitative comparison.}
Figure~\ref{fig:visual_comparison} presents three selected DIV2K and RealSR cases.
For transparent plastic, FiDeSR and AdcSR introduce dense crosshatched texture, whereas SPGSR better preserves the broad folds visible in HR.
For the eye, FiDeSR accentuates eyelashes and VOSR-1.4B alters surrounding strokes, whereas SPGSR more closely follows the reference contours.
For the statue drapery, SPGSR retains the broad fold geometry with less spurious fine texture than VOSR-1.4B and FiDeSR.

\subsection{Ablation Studies}
\label{sec:ablations}
Tables~\ref{tab:sr_ablation} and~\ref{tab:sr_state} compare conditioning, adversarial supervision, refinement, capacity, and state/prediction choices under the matched recipe.

\begin{table}[t]
\centering
\captionsetup{skip=4pt}
\caption{Ablations of DINO conditioning and supervision, pixel-space refinement, and transformer depth on RealSR.}
\label{tab:sr_ablation}
\footnotesize
\setlength{\tabcolsep}{2pt}
\begin{tabular}{lrrrrr}
\toprule
Configuration & Gen. (M) & DISTS$\downarrow$ & CLIPIQA$\uparrow$ & MUSIQ$\uparrow$ & MANIQA$\uparrow$ \\
\midrule
\multicolumn{6}{l}{\textit{DINO conditioning and supervision}} \\
w/o DINO cond. & 23.567 & 0.2076 & 0.6586 & 68.75 & 0.6363 \\
Single layer 1 & 31.748 & 0.2056 & 0.6570 & 68.15 & 0.6255 \\
Single layer 2 & 31.748 & 0.2037 & 0.6647 & 68.23 & 0.6297 \\
Single layer 4 & 31.748 & 0.2010 & 0.6677 & 68.72 & 0.6429 \\
Single layer 6 & 31.748 & 0.1997 & 0.7003 & 69.14 & 0.6511 \\
Single layer 8 & 31.748 & 0.1970 & 0.6980 & 69.35 & 0.6582 \\
Single layer 10 & 31.748 & 0.1971 & 0.6915 & 69.19 & 0.6567 \\
Uniform fusion & 31.748 & 0.1961 & 0.6764 & 68.82 & 0.6494 \\
w/o routing loss & 31.799 & 0.1951 & 0.6874 & 68.66 & 0.6478 \\
w/o DINO-GAN & 31.799 & 0.1924 & 0.5839 & 60.74 & 0.5253 \\
\textbf{SPGSR} & 31.799 & 0.2001 & 0.6765 & 70.05 & 0.6639 \\
\midrule
\multicolumn{6}{l}{\textit{Pixel-space refinement}} \\
w/o PixelRRB & 31.498 & 0.2007 & 0.6500 & 69.61 & 0.6635 \\
FFN-matched & 31.803 & 0.1989 & 0.6470 & 69.62 & 0.6640 \\
\midrule
\multicolumn{6}{l}{\textit{Transformer depth (width 384, six heads)}} \\
8 blocks & 22.325 & 0.2025 & 0.6729 & 70.27 & 0.6628 \\
16 blocks & 41.273 & 0.2003 & 0.6527 & 69.79 & 0.6627 \\
\bottomrule
\end{tabular}
\end{table}

\paragraph{DINO conditioning and routing.}
Removing DINO conditioning lowers MUSIQ/MANIQA to 68.75/0.6363.
This control also reduces generator capacity and changes loss normalization.
Single-layer controls retain six-layer feature and GAN supervision.
SPGSR has higher MUSIQ/MANIQA than every single-layer control.

Uniform fusion assigns equal weights to the six selected DINO layers.
The w/o routing loss control retains the trainable router but removes the routing loss, learning the fusion weights through the remaining restoration objectives.
Uniform fusion and this control yield MUSIQ 68.82 and 68.66, respectively, versus 70.05 for SPGSR.
Removing the routing loss also removes its contribution to adaptive normalization.

\paragraph{DINO adversarial supervision.}
This control uses the same Stage 1 EMA initialization and retains all other Stage 2 objectives.
Removing DINO-GAN reduces MUSIQ from 70.05 to 60.74 and MANIQA from 0.6639 to 0.5253.
PSNR, SSIM, LPIPS, DISTS, and FID all improve.

\paragraph{Pixel-space refinement.}
PixelRRB adds $0.30\mathrm{M}$ generator parameters.
The FFN-matched control removes PixelRRB and widens the feed-forward layers, keeping approximately the same generator parameter count as SPGSR.
SPGSR improves CLIPIQA from 0.6470 to 0.6765 and MUSIQ from 69.62 to 70.05, while DISTS and MANIQA do not improve.

\paragraph{Generator capacity.}
The default generator uses 12 Transformer blocks.
Reducing this number to 8 removes 29.8\% of generator parameters while increasing MUSIQ by 0.22 on RealSR.
Increasing it to 16 slightly improves PSNR but worsens DISTS, CLIPIQA, and MUSIQ.

\begin{table}[t]
\centering
\caption{State and prediction controls on RealSR. Fixed/random refers to Stage 1 time sampling. Stage 2 always uses fixed time.}
\label{tab:sr_state}
\footnotesize
\setlength{\tabcolsep}{2.8pt}
\begin{tabular}{lrrrr}
\toprule
Prediction / Stage 1 & DISTS$\downarrow$ & CLIPIQA$\uparrow$ & MUSIQ$\uparrow$ & FID$\downarrow$ \\
\midrule
RGB / fixed & 0.1993 & 0.6444 & 69.52 & 108.58 \\
residual / fixed & 0.1981 & 0.6591 & 69.35 & 107.18 \\
\textbf{SPGSR}: residual / random & 0.2001 & 0.6765 & 70.05 & 113.68 \\
\bottomrule
\end{tabular}
\end{table}

\paragraph{State and prediction controls.}
Table~\ref{tab:sr_state} compares direct RGB and residual prediction with different Stage 1 time sampling.
SPGSR improves CLIPIQA and MUSIQ over fixed-time residual training, but worsens DISTS and FID.
Direct RGB prediction remains competitive.
The supplement provides all nine metrics for these controls.

\section{Conclusion}
\label{sec:conclusion}
We presented SPGSR, a compact vision-only framework for one-step super-resolution with a 31.8M-parameter pixel-space generator.
By combining LR-anchored residual prediction, lightweight local refinement, and pretrained visual guidance, SPGSR achieves competitive restoration quality across three benchmarks without relying on pretrained T2I generators or SR teacher distillation.
Its compact design also enables low inference latency and peak GPU memory usage.
Compared with VOSR-0.5B and VOSR-1.4B, SPGSR improves PSNR, SSIM, LPIPS, and DISTS on all three benchmarks, with 93.5\% and 97.7\% fewer generator parameters and $2.80\times$ and $3.35\times$ GPU-pipeline speedups under FP16 inference, respectively.
These results suggest that competitive vision-only one-step SR can be achieved with substantially smaller generators, without sacrificing the benefits of generative restoration.

{\small
    \bibliographystyle{ieeenat_fullname}
    \bibliography{main}
}

\clearpage
\setcounter{section}{0}
\setcounter{subsection}{0}
\setcounter{table}{0}
\setcounter{figure}{0}
\setcounter{equation}{0}
\renewcommand{\thesection}{S\arabic{section}}
\renewcommand{\thetable}{S\arabic{table}}
\renewcommand{\thefigure}{S\arabic{figure}}
\renewcommand{\theequation}{S\arabic{equation}}
\renewcommand{\theHsection}{supp.\arabic{section}}
\renewcommand{\theHtable}{supp.\arabic{table}}
\renewcommand{\theHfigure}{supp.\arabic{figure}}
\renewcommand{\theHequation}{supp.\arabic{equation}}
\twocolumn[{\centering\Large\bfseries Supplementary Material\par\vspace{1em}}]
\raggedbottom
\section{Implementation and Training Details}

PixelRRB maps the concatenated LR image (upsampled) and transformer residual from six to 64 channels using a convolution and ReLU, followed by four residual blocks, each with two convolutions, an intervening ReLU and an identity skip connection.
A zero-initialized output convolution produces a three-channel correction added to the transformer residual.
All convolutions use $3\times3$ kernels, stride 1 and padding 1.

DINOv2-S uses $448\times448$ inputs and zero-based layers $\{1,2,4,6,8,10\}$.
Predictions are clipped to $[-1,1]$ for DINO feature and LPIPS losses, but not for flow loss.
Discriminator heads use spectral normalization and real/fake targets of $0.8/0$.

We use AdamW for the generator in both stages and the discriminator in Stage 2, with $\beta=(0.9,0.95)$, weight decay 0.01 and gradient-norm clipping at 1.
The generator's EMA decay is 0.9999.

\paragraph{Training-budget accounting.}
In Figure 1, training pixels count image exposure during generator pretraining and SR adaptation.
They exclude separate teacher or discriminator forward passes and pretraining of reused visual encoders and VAEs.
SD pretraining and OSEDiff/PiSA-SR adaptation budgets follow VOSR Figure 1, and VOSR's own budgets follow its Appendix A.3.
FiDeSR and AdcSR budgets use their paper-reported training settings, including inherited teacher training for AdcSR.
CODSR's adaptation budget is estimated from its public two-stage launchers.

\paragraph{Inference measurements.}
GPU-pipeline latency is measured as median CUDA-synchronized wall time, excluding disk I/O and model loading.
Memory is the maximum PyTorch peak allocation across three test inputs after warm-up.

\section{Extended Ablations}

Tables~\ref{tab:supp_mechanism}, \ref{tab:supp_depth} and~\ref{tab:supp_state} provide all nine metrics for the RealSR ablations discussed in the main paper.

In Table~\ref{tab:supp_cross_dataset}, SPGSR has higher MUSIQ and MANIQA than all four controls on both datasets.
The routing controls have higher PSNR and SSIM, while the refinement controls have lower NIQE and FID.

\onecolumn
\par\medskip
\noindent\begin{minipage}{\textwidth}
\centering
\captionsetup{hypcap=false}
\captionof{table}{DINO-conditioning, adversarial-supervision and PixelRRB ablations on RealSR.}
\label{tab:supp_mechanism}
\footnotesize
\setlength{\tabcolsep}{2.8pt}
\begin{tabular}{lrrrrrrrrr}
\toprule
Configuration & PSNR$\uparrow$ & SSIM$\uparrow$ & LPIPS$\downarrow$ & DISTS$\downarrow$ & CLIPIQA$\uparrow$ & NIQE$\downarrow$ & MUSIQ$\uparrow$ & MANIQA$\uparrow$ & FID$\downarrow$ \\
\midrule
w/o DINO conditioning & 26.19 & 0.7502 & 0.2707 & 0.2076 & 0.6586 & 5.0905 & 68.75 & 0.6363 & 132.17 \\
Single layer 1 & 26.37 & 0.7573 & 0.2633 & 0.2056 & 0.6570 & 5.1127 & 68.15 & 0.6255 & 130.23 \\
Single layer 2 & 26.40 & 0.7550 & 0.2641 & 0.2037 & 0.6647 & 5.2680 & 68.23 & 0.6297 & 129.76 \\
Single layer 4 & 26.27 & 0.7539 & 0.2614 & 0.2010 & 0.6677 & 4.9611 & 68.72 & 0.6429 & 125.18 \\
Single layer 6 & 26.21 & 0.7541 & 0.2603 & 0.1997 & 0.7003 & 5.1668 & 69.14 & 0.6511 & 120.71 \\
Single layer 8 & 26.18 & 0.7523 & 0.2597 & 0.1970 & 0.6980 & 5.1014 & 69.35 & 0.6582 & 114.54 \\
Single layer 10 & 26.17 & 0.7547 & 0.2581 & 0.1971 & 0.6915 & 5.0242 & 69.19 & 0.6567 & 111.99 \\
Uniform fusion & 26.41 & 0.7586 & 0.2565 & 0.1961 & 0.6764 & 5.1558 & 68.82 & 0.6494 & 111.84 \\
w/o routing loss & 26.40 & 0.7590 & 0.2578 & 0.1951 & 0.6874 & 5.1367 & 68.66 & 0.6478 & 113.12 \\
w/o DINO-GAN & 27.54 & 0.7815 & 0.2397 & 0.1924 & 0.5839 & 5.9752 & 60.74 & 0.5253 & 109.01 \\
w/o PixelRRB & 26.07 & 0.7445 & 0.2659 & 0.2007 & 0.6500 & 5.0128 & 69.61 & 0.6635 & 112.03 \\
FFN-matched & 26.10 & 0.7454 & 0.2649 & 0.1989 & 0.6470 & 5.0371 & 69.62 & 0.6640 & 112.58 \\
\textbf{SPGSR (12 blocks)} & 26.12 & 0.7507 & 0.2638 & 0.2001 & 0.6765 & 5.1008 & 70.05 & 0.6639 & 113.68 \\
\bottomrule
\end{tabular}
\end{minipage}\par

\par\medskip
\noindent\begin{minipage}{\textwidth}
\centering
\captionsetup{hypcap=false}
\captionof{table}{Transformer-depth ablations on RealSR, with width and head count fixed.}
\label{tab:supp_depth}
\footnotesize
\setlength{\tabcolsep}{2.8pt}
\begin{tabular}{lrrrrrrrrr}
\toprule
Configuration & PSNR$\uparrow$ & SSIM$\uparrow$ & LPIPS$\downarrow$ & DISTS$\downarrow$ & CLIPIQA$\uparrow$ & NIQE$\downarrow$ & MUSIQ$\uparrow$ & MANIQA$\uparrow$ & FID$\downarrow$ \\
\midrule
8 blocks & 26.12 & 0.7484 & 0.2643 & 0.2025 & 0.6729 & 5.0750 & 70.27 & 0.6628 & 112.91 \\
\textbf{SPGSR (12 blocks)} & 26.12 & 0.7507 & 0.2638 & 0.2001 & 0.6765 & 5.1008 & 70.05 & 0.6639 & 113.68 \\
16 blocks & 26.15 & 0.7523 & 0.2641 & 0.2003 & 0.6527 & 5.1172 & 69.79 & 0.6627 & 112.82 \\
\bottomrule
\end{tabular}
\end{minipage}\par

\par\medskip
\noindent\begin{minipage}{\textwidth}
\centering
\captionsetup{hypcap=false}
\captionof{table}{Prediction type and Stage 1 time-sampling ablations on RealSR.}
\label{tab:supp_state}
\footnotesize
\setlength{\tabcolsep}{2.8pt}
\begin{tabular}{lrrrrrrrrr}
\toprule
Configuration & PSNR$\uparrow$ & SSIM$\uparrow$ & LPIPS$\downarrow$ & DISTS$\downarrow$ & CLIPIQA$\uparrow$ & NIQE$\downarrow$ & MUSIQ$\uparrow$ & MANIQA$\uparrow$ & FID$\downarrow$ \\
\midrule
direct RGB, fixed time & 26.18 & 0.7509 & 0.2638 & 0.1993 & 0.6444 & 5.1234 & 69.52 & 0.6639 & 108.58 \\
residual, fixed time & 26.20 & 0.7503 & 0.2635 & 0.1981 & 0.6591 & 5.0198 & 69.35 & 0.6594 & 107.18 \\
\textbf{SPGSR}: residual, random time & 26.12 & 0.7507 & 0.2638 & 0.2001 & 0.6765 & 5.1008 & 70.05 & 0.6639 & 113.68 \\
\bottomrule
\end{tabular}
\end{minipage}\par

\par\medskip
\noindent\begin{minipage}{\textwidth}
\centering
\captionsetup{hypcap=false}
\captionof{table}{Conditioning and refinement ablations on DRealSR and DIV2K.}
\label{tab:supp_cross_dataset}
\footnotesize
\setlength{\tabcolsep}{2.8pt}
\begin{tabular}{lrrrrrrrrr}
\toprule
Configuration & PSNR$\uparrow$ & SSIM$\uparrow$ & LPIPS$\downarrow$ & DISTS$\downarrow$ & CLIPIQA$\uparrow$ & NIQE$\downarrow$ & MUSIQ$\uparrow$ & MANIQA$\uparrow$ & FID$\downarrow$ \\
\midrule
\multicolumn{10}{l}{\textit{DRealSR (93 pairs)}} \\
Uniform fusion & 28.75 & 0.7791 & 0.3063 & 0.2154 & 0.7186 & 5.6698 & 65.28 & 0.6112 & 140.17 \\
w/o routing loss & 28.75 & 0.7798 & 0.3057 & 0.2170 & 0.7259 & 5.7820 & 65.53 & 0.6118 & 140.36 \\
w/o PixelRRB & 28.46 & 0.7628 & 0.3157 & 0.2200 & 0.6918 & 5.5042 & 66.08 & 0.6213 & 136.67 \\
FFN-matched & 28.53 & 0.7654 & 0.3117 & 0.2178 & 0.6933 & 5.5486 & 65.92 & 0.6191 & 136.68 \\
\textbf{SPGSR} & 28.53 & 0.7729 & 0.3124 & 0.2193 & 0.7079 & 5.7070 & 66.11 & 0.6237 & 138.36 \\
\midrule
\multicolumn{10}{l}{\textit{DIV2K (3,000 pairs)}} \\
Uniform fusion & 24.61 & 0.6211 & 0.2711 & 0.1777 & 0.7079 & 4.6686 & 68.75 & 0.6311 & 21.84 \\
w/o routing loss & 24.61 & 0.6217 & 0.2715 & 0.1775 & 0.7112 & 4.6268 & 68.84 & 0.6291 & 21.66 \\
w/o PixelRRB & 24.35 & 0.6076 & 0.2756 & 0.1794 & 0.6454 & 4.4304 & 68.99 & 0.6398 & 21.57 \\
FFN-matched & 24.36 & 0.6085 & 0.2751 & 0.1792 & 0.6442 & 4.4678 & 69.00 & 0.6391 & 21.41 \\
\textbf{SPGSR} & 24.38 & 0.6140 & 0.2752 & 0.1795 & 0.7051 & 4.5717 & 69.54 & 0.6444 & 21.87 \\
\bottomrule
\end{tabular}
\end{minipage}\par

\end{document}